\documentclass[letterpaper, 10 pt, journal, twoside]{IEEEtran}

\IEEEoverridecommandlockouts                              

\usepackage{bm}
\usepackage{float}
\usepackage[pdftex]{graphicx}
\usepackage{tabularx}
\usepackage{verbatim}
\usepackage{color}
\usepackage{xparse}
\usepackage{hyperref}
\usepackage{xmpmulti}
\usepackage{transparent}
\usepackage{fancyhdr}
\usepackage{cite}
\usepackage{rotating}

\usepackage[ruled, lined, linesnumbered, commentsnumbered, longend]{algorithm2e}

\usepackage{booktabs}
\usepackage{caption}
\usepackage{subcaption}
\usepackage{tikz}
\usepackage{hyperref}
\usepackage{wrapfig}
\usepackage{setspace}
\usepackage{graphicx}
\usepackage[all]{xy}
\usepackage{verbatim}
\usepackage{float}
\usepackage{mathrsfs}  
\usepackage{bm}
\usepackage{multirow}
\usepackage{color,soul}
  
\usepackage{amssymb,amsmath,graphicx}
  
\usepackage{amsthm}
\usepackage[compact]{titlesec}
\usepackage{diagbox}
\usepackage{array}
\usepackage{titlesec}
\usepackage{units}
\usepackage[export]{adjustbox}

\usepackage[letterpaper, left=0.7in, right=0.7in, bottom=0.6in, top=0.8in]{geometry}
\let\oldIEEEkeywords\IEEEkeywords
\def\IEEEkeywords{\oldIEEEkeywords\normalfont\bfseries\ignorespaces}

\newtheorem{example}{Example}
\newtheorem{problem}{Problem}
                       
\newtheorem{definition}{Definition}

\usepackage{mathtools}
\mathtoolsset{showonlyrefs}

\title{\LARGE \bf
Weighted Graph-Based Signal Temporal Logic Inference Using Neural Networks
}

\author{Nasim Baharisangari$^{1}$, Kazuma Hirota$^{2}$, Ruixuan Yan$^{3}$, Agung Julius$^{3}$, and Zhe Xu$^{1}$
\thanks{This work was partially funded by the Defense Advanced Research Projects Agency (DARPA) under Contract No. HR001120C0032 and the NSF grant 1936578. (Corresponding author: Zhe Xu.)}
\thanks{$^{1}$Nasim Baharisangari and Zhe Xu are with the School for Engineering of Matter, Transport and Energy, Arizona State University, Tempe, AZ 85287. $^2$Kazuma Hirota is with Walker Department of Mechanical Engineering, The University of Texas at Austin, Austin, TX 78712. $^{3}$Ruixuan Yan and Agung Julius are with the Department of Electrical, Computer, and Systems Engineering, Rensselaer Polytechnic Institute, 110 8th St, Troy, NY 12180. {\tt\small $\{$nbaharis, xzhe1$\}$@asu.edu, kazhirota7@utexas.edu, yanr5@rpi.edu, agung@ecse.rpi.edu }}}

\titleformat{\section}{\centering\normalfont\scshape}{\thesection}{0em}{.~}
\titleformat{\paragraph}[runin]{\bfseries}{}{-2em}{}[:]

\DeclareMathAlphabet{\mathpzc}{OT1}{pzc}{m}{it}

\newcommand{\subformulas}[1]{\ensuremath{\mathit{subf(#1)}}}
\DeclareMathOperator{\ltrue}{\mathit{True}}

\DeclareMathOperator{\limplies}{\rightarrow}

\DeclareMathOperator{\leventually}{\mathbf{F}}
\DeclareMathOperator{\lglobally}{\mathbf{G}}

\newcommand{\abs}[1]{\ensuremath{|#1|}}

\newcommand{\real}{\ensuremath{\mathbb{R}}}

\newcommand{\formula}{\ensuremath{\phi}}

\newcommand{\formulaG}{\ensuremath{\phi_\mathcal{G}}}

\newcommand{\formulaGSt}{\ensuremath{\phi^1_{\mathcal{G}}}}

\newcommand{\formulaGNd}{\ensuremath{\phi^2_{\mathcal{G}}}}

\newcommand{\Gexists}{\ensuremath{\boldsymbol{\exists_{\mathcal{G}}}}}

\newcommand{\Gforall}{\ensuremath{{\boldsymbol{\forall_{\mathcal{G}}}}}}

\newcommand{\node}{\ensuremath{v}}

\newcommand{\Gennode}{\ensuremath{v}}

\newcommand{\hatnode}{\ensuremath{\hat{v}}}

\newcommand{\nodeNo}{\ensuremath{n_N}}

\newcommand{\nodeSet}{\ensuremath{V}}

\newcommand{\edgeNo}{\ensuremath{n_E}}

\newcommand{\graphG}{\ensuremath{G}}

\NewDocumentCommand{\graphTraj}{O{}}{\ensuremath{g_{#1}}}

\newcommand{\nodeLabelSet}{\ensuremath{\mathcal{Z}}}

\newcommand{\robustnessG}{\ensuremath{r}} 

\newcommand{\counterj}{\ensuremath{j}}

\newcommand{\dimcounter}{\ensuremath{d}}

\newcommand{\counterk}{\ensuremath{k}}

\newcommand{\timeIndex}{\ensuremath{k}}
\newcommand{\timeIndexSt}{\ensuremath{\timeIndex_1}}
\newcommand{\timeIndexNd}{\ensuremath{\timeIndex_2}}

\newcommand{\genPair}[2]{\ensuremath{(\node,\timeIndex)}}

\newcommand{\timeInterval}{\ensuremath{I=[\timeIndexSt,\timeIndexNd]}}

\newcommand{\SampleStyle}[1]{\ensuremath{ \mathcal{#1} }}

\NewDocumentCommand{\Sample}{O{}}{\ensuremath{\SampleStyle{D}^{#1}}}

\NewDocumentCommand{\GSSample}{O{}}{\ensuremath{\SampleStyle{D_G}^{#1}}}

\NewDocumentCommand{\class}{O{}}{\ensuremath{l}_{#1}} 
\newcommand{\posclass}{\ensuremath{+1}}
\newcommand{\negclass}{\ensuremath{-1}}

\NewDocumentCommand{\robustness}{O{}}{\ensuremath{r_{#1}}} 

\SetKwInput{Hyperparameter}{Hyperparameter}
\SetKwInput{Input}{Input}
\SetKwInput{Output}{Output}
\SetKwBlock{Loop}{loop}{end}

\SetKwInOut{Parameter}{Parameter}
\SetKwFunction{AlgoLearnLTL}{\text{Learn-LTL}}
\newcommand{\DTree}{\text{Decision-tree}}
\SetKwFunction{StopCriteria}{\text{Stopping-Criteria}}
\SetKwFunction{AlgoDTree}{\DTree}
\newcommand{\SplitSample}{\text{Split-sample}}
\SetKwFunction{AlgoSplitSample}{\SplitSample}
\newcommand{\Optimize}{\text{Infer-formula}}
\SetKwFunction{AlgoOptimize}{\Optimize}
\newcommand{\OptimizeMaxedSize}{\text{Infer-formula-max}}
\SetKwFunction{AlgoOptimizeMaxedSize}{\OptimizeMaxedSize}
\newcommand{\OptimizeSufficientScore}{\text{Infer-formula-min}}
\SetKwFunction{AlgoOptimizeSufficientScore}{\OptimizeSufficientScore}

\newcommand{\I}{\ensuremath{I}}
\newcommand{\counteri}{\ensuremath{i}}

\newcommand{\counterl}{\ensuremath{l}}
\NewDocumentCommand{\weight}{O{}}{\ensuremath{{w}_{#1}}}
\newcommand{\weightSt}{\ensuremath{{w}_1}}
\newcommand{\weightNd}{\ensuremath{{w}_2}}
\newcommand{\Capweight}{\ensuremath{\mathbf{\Omega}}}
\newcommand{\supWeight}{\ensuremath{\rm{w}}}
\newcommand{\GCapweight}{\ensuremath{\boldsymbol{\mathcal{W}}}}
\NewDocumentCommand{\normalweight}{o}{\ensuremath{\overline{\weight}_{#1}}}
\newcommand{\wSat}[3]{\ensuremath{r^{\supWeight}(#1,#2,#3)}}
\newcommand{\actFunAnd}{\ensuremath{\otimes^{\land}}}
\newcommand{\actFunOr}{\ensuremath{\otimes^{\lor}}}
\newcommand{\actFunGlo}{\ensuremath{\otimes^{\lglobally}}}
\newcommand{\actFunEve}{\ensuremath{\otimes^{\leventually}}}

\newcommand{\actFunGexists}{\ensuremath{\otimes^{\Gexists}}}

\newcommand{\actFunGforall}{\ensuremath{\otimes^{\Gforall}}}

\newcommand{\wSTL}{\ensuremath{\Tilde{\phi}}}

\NewDocumentCommand{\GwSTL}{O{}}{\ensuremath{\textsuperscript{#1}\Tilde{\phi}_{\mathcal{G}}}}

\NewDocumentCommand{\wSTLstyle}{O{}O{}O{}}{\ensuremath{\textsuperscript{#1}\wSTL^{#2}_{#3}}} 
\NewDocumentCommand{\intervalStyle}{mmO{}O{}}{\ensuremath{\{{#1},{#2}\}_{#3}^{#4}}}

\newcommand{\counteriequal}{\ensuremath{i=1,2}}

\newcommand{\counteriin}{\ensuremath{i\in{\timeIndex+\I}}}
\NewDocumentCommand{\wfinally}{O{}O{}}{\ensuremath{\leventually^{#1}_{#2}}}
\NewDocumentCommand{\walways}{O{}O{}}{\ensuremath{\lglobally^{#1}_{#2}}}

\NewDocumentCommand{\wimplies}{O{}O{}}{\ensuremath{\limplies^{#1}_{#2}}}

\newcommand{\lSigma}[3]{\ensuremath{{\sum}^{#1}_{#2}}{#3}}

\newcommand{\lSigmaTop}[3]{\ensuremath{{\sum\limits^{#1}_{#2}{#3}}}}

\newcommand{\promVar}{\ensuremath{\sigma}}

\newcommand{\neighbor}{\ensuremath{\bigcirc_{\mathrm{N}}}}

\newcommand{\neighborI}{\ensuremath{\abs{\bigcirc_{\mathrm{N}}\node}}}

\newcommand{\siBool}{\ensuremath{s_{\iprim}=e^{-\frac{r_{\iprim}}{\promVar}}/\lSigma{2}{\counterl=1}{{e^{-\frac{r_{\counterl}}{\promVar}}}}}}

\newcommand{\siTemp}{\ensuremath{s_{\iprim}=e^{-\frac{r_{\iprim}}{\promVar}}/\lSigma{\timeIndexNd}{\counterl=\timeIndexSt}{{e^{-\frac{r_{\counterl}}{\promVar}}}}}}

\newcommand{\siGraph}{\ensuremath{s_{\iprim}=e^{-\frac{r_{\iprim}}{\promVar}}/\lSigma{\neighborI}{\counterl=1}{{e^{-\frac{r_{\counterl}}{\promVar}}}}}}

\newcommand{\ActiveS}{\ensuremath{s}}

\NewDocumentCommand{\FormulaStruct}{O{}}{\ensuremath{\mathpzc{F}_{\rm{#1}}}}

\newcommand{\PosSet}{\ensuremath{\mathcal{D}_P}}

\newcommand{\NegSet}{\ensuremath{\mathcal{D}_N}}

\NewDocumentCommand{\TotSet}{O{}}{\ensuremath{\mathcal{D}_C}^{#1}}

\newcommand{\PosSetSize}{\ensuremath{\Tilde{p}}}

\newcommand{\NegSetSize}{\ensuremath{\Tilde{n}}}

\newcommand{\TotSetSize}{\ensuremath{N_{\Sample}}}

\newcommand{\SubFormNo}{\ensuremath{\mathcal{J}}}

\newcommand{\Setlabel}{\ensuremath{l_{\counteri}}}

\newcommand{\tuningParam}{\ensuremath{\eta}}

\newcommand{\wSatTwo}[2]{\ensuremath{r^{\supWeight}(#1,#2)}}

\newcommand{\LossFun}{\ensuremath{J}}

\newcommand{\STLCo}{\ensuremath{\mathbf{a_S}}}

\newcommand{\STLbias}{\ensuremath{{c}}}

\newcommand{\batchSize}{\ensuremath{K}}

\newcommand{\STLf}{\ensuremath{{f}}}

\NewDocumentCommand{\traceTime}{O{}O{}}{\ensuremath{\zeta^{#2}(#1)}}

\NewDocumentCommand{\traceDim}{O{}}{\ensuremath{\zeta_{#1}}}

\newcommand{\GforallCo}{\ensuremath{q_{\Gforall}}}

\newcommand{\GexistsCo}{\ensuremath{q_{\Gexists}}}

\newcommand{\globallyCo}{\ensuremath{q_{\lglobally}}}

\newcommand{\eventuallyCo}{\ensuremath{q_{\leventually}}}

\newcommand{\pTemp}{\ensuremath{P_{\textrm{temporal}}}}

\newcommand{\pGL}{\ensuremath{P_{\textrm{\GL}}}}

\NewDocumentCommand{\actFun}{O{}}{\ensuremath{\otimes^{#1}}}

\newcommand{\nodeVal}{node value}

\newcommand{\GL}{GL}

\newcommand{\GSTL}{GSTL}

\newcommand{\wGSTL}{w-GSTL}

\newcommand{\wtsGLNN}{w-GSTL-NN}
 
\newcommand{\StructName}{flexible \wGSTL{} formula structure}

\newcommand{\back}{\ensuremath{F_{\textrm{backward}}}}

\newcommand{\forward}{\ensuremath{F_{\textrm{forward}}}}

\newcommand{\param}{\ensuremath{Parameters}}

\newcommand{\iprim}{\ensuremath{m}}

\newcommand{\accu}{\ensuremath{AC}}

\newcommand{\Predicted}{\ensuremath{{\boldsymbol{l}}^{\rm{predicted}}}}

\begin{document}

\maketitle
\thispagestyle{empty}
\pagestyle{empty}

\begin{abstract}\label{abstract}Extracting spatial-temporal knowledge from data is useful in many applications. It is important that the obtained knowledge is human-interpretable and amenable to formal analysis. In this paper, we propose a method that trains neural networks to learn  spatial-temporal properties in the form of \textit{weighted graph-based signal temporal logic} (\wGSTL{}) formulas. For learning  \wGSTL{} formulas, we introduce a \textit{flexible} \wGSTL{} \textit{formula structure} in which the user's preference can be applied in the inferred \wGSTL{} formulas. In the proposed framework, each neuron of the neural networks corresponds to a subformula in a \textit{flexible} \wGSTL{} \textit{formula structure}. We initially train a neural network to learn the \wGSTL{} operators, and then train a second neural network to learn the parameters in a \textit{flexible} \wGSTL{} \textit{formula structure}. We use a COVID-19 dataset and a rain prediction dataset to evaluate the performance of the proposed framework and algorithms. We compare the performance of the proposed framework with four baseline classification methods including K-nearest neighbors, decision trees, support vector machine, and artificial neural networks. The classification accuracy obtained by the proposed framework is comparable with the baseline classification methods.

\end{abstract}


\section{Introduction}\label{Se.Intro}
Learning spatial-temporal properties from data is useful in many applications, especially where the data is based on an underlying graph structure.
It is preferable that the learned properties are interpretable for humans and amenable to formal analysis \cite{Seshia2016}. Various logics have been introduced to express and analyze spatial-temporal properties \cite{Xu2019}\cite{Haghighi2016}\cite{Djeumou2020}, and temporal logics are one of the major groups. Temporal logics, which are categorized as formal languages \cite{Hopcroft1979IntroductionTA}, demonstrate both the interpretability  and being amenable to formal analysis; thus, temporal logics are used to express the temporal and logical properties of systems in a human-interpretable way {\cite{Linard}}. In addition, \textit{graph-based logic} (\GL), which is used to express spatial properties, is understandable for humans and preserves the rigorous aspect of formal logics. \par

Besides interpretability and being amenable to formal analysis, efficiency and expressiveness \cite{Guhring2020} are also important when learning spatial-temporal properties from data \cite{Liu2020}. One approach to increase the efficiency of the learning task is to integrate neural networks into the process \cite{Seo2017} \cite{Wu2019}.
We can expand the capacity of the learning task to handle more complex spatial-temporal datasets by combining the distinct advantages of temporal logics and graph-based logics, and neural networks.\par
{\textit{Signal temporal logic} (STL) is one type of temporal logics, which deals with real-valued data over real-time domain \cite{Donze}. In this paper, we combine STL and GL to introduce \textit{graph-based signal temporal logic} (\GSTL{}) formulas to express spatial-temporal properties. Furthermore, we assign \textit{importance weights} to the subformulas of a GSTL formula and name it \textit{weighted graph-based signal temporal logic} (w-GSTL), where each importance weight quantifies the importance of a subformula of a w-GSTL formula. }
\paragraph{Contributions}
In this paper, we propose a methodology that trains neural networks to learn spatial-temporal properties from data in the form of {\wGSTL{}} formulas. The contributions of this paper are as follow: (a) we introduce a \textit{flexible} \wGSTL{} \textit{formula structure} that allows the user's preference to be applied in the inferred \wGSTL{} formula. In this structure, the \wGSTL{} operators are free to be inferred; (b)
we propose a framework and algorithms to learn \wGSTL{} formulas from data using neural networks. In the proposed framework and algorithms, neurons of the neural networks represent the quantitative satisfaction of the \wGSTL{} operators and Boolean connectives. For a given \textit{flexible} \wGSTL{} \textit{formula structure}, we first construct and train a neural network to learn \wGSTL{} operators through back-propagation; then, we construct and train another neural network to learn the parameters of the \textit{flexible} \wGSTL{} \textit{formula structure} through back-propagation. We evaluate the performance of the proposed framework and algorithms by exploiting real-life examples: predicting COVID-19 lockdown measure in a geographical region in Italy, and rainfall prediction in a geographical region in Australia. The obtained results show that the proposed method achieves comparable classification accuracy with comparison with four baseline classification methods including K-nearest neighbors (KNN), decision trees (DT), {support vector machine (SVM)}, and  artificial neural networks (ANN), while the interpretability has been improved with the learned \wGSTL{} formulas.

\subsection{Related Work}

Recently, learning spatial-temporal properties from data has been employed in different applications such as swarm robotics \cite{Djeumou2020}, etc. Different methods have been adopted to carry out this learning task. Many researchers have developed different logics to learn spatial-temporal properties from data. For example, Xu \textit{et. al.} \cite{Xu2019} introduce \textit{graph temporal logic} (GTL), or Liu \textit{et. al.} \cite{Liu2020} introduce \textit{graph-based spatial temporal logic} (GSTL).
Many other researchers propose frameworks to conduct the learning tasks based on neural networks. For instance, Wu \textit{et. al.} \cite{Wu2019} develop a CNN (\textit{convolution neural network})-based method and name it \textit{Graph WaveNet}.
and, Ziat \textit{et. al.} \cite{Ziat2017} introduce \textit{Spatio-Temporal Neural Network} (STNN).
The proposed approach in this paper benefits from advantages of both the formal logics and neural networks: human-interpretability and efficiency.\par

Moreover, combining temporal logic and neural networks to carry out learning tasks has been gaining attention \cite{Yan2021} \cite{Serafini2016}\cite{Riegel2020}. One way to realize this combination is through connecting the temporal operators and Boolean connectives to the activation functions in neural networks \cite{Riegel2020}. Most of the standard algorithms used  to conduct logic inference solve a non-convex optimization problem to find parameters in the formula, where the loss function of a neural network is not differentiable  with respect to the parameters at every point. In \cite{Yan2021}, Yan \textit{et. al.} propose a loss function that addresses the differentiability issue. In addition, the proposed frameworks in \cite{Yan2021}, \cite{Riegel2020}, and \cite{Serafini2016} do not extract spatial-temporal properties from data. \par

\section{Preliminaries}\label{Sec:Prelim}

\paragraph{Graph} We denote a graph by $\graphG=(\nodeSet,E)$, where $\nodeSet=\{\Gennode_1,\Gennode_2,...,\Gennode_{\nodeNo}\}$ is a finite set of nodes,  $E=\{e_1,e_2,...,e_{\edgeNo}\}$ is a finite set of edges, and $\nodeNo,\edgeNo\in\mathbb{N}=\{1,2,...\}$. We also denote a set of (possibly infinite)  \nodeVal{s} by $\nodeLabelSet$, where $\nodeLabelSet\subseteq\real$. 

\vspace{-3pt}

\paragraph{Graph-based trajectory} 
We define a finite $\dimcounter$-dimensional \textit{graph-based trajectory} $\graphTraj:\nodeSet\times\mathbb{T}\limplies\nodeLabelSet^{d}$ that assigns a node value for each node $\node$ at time-step $\timeIndex\in\mathbb{T}=\{0,1,2,...,J\}$, where $T$ is a finite discrete time domain and $J\in\mathbb{N}$. We also denote the value of the $\dimcounter$-th dimension of the graph-based trajectory $\graphTraj$ at time-step $\timeIndex$ and node $\node$ by $\graphTraj^{\dimcounter}(\node,\timeIndex)$. A time interval is denoted by $\timeInterval=\{\timeIndex'|\timeIndexSt\leq{\timeIndexNd},~\timeIndexSt,\timeIndexNd\in\mathbb{T}\}$, and $\timeIndex+I$ denotes the time interval $[\timeIndex+\timeIndexSt, \timeIndex+\timeIndexNd]$.
\vspace{-2pt}
\section{Weighted Graph-Based Signal Temporal Logic}\label{Sec:Prob.Form}

In this section, we introduce \textit{weighted graph-based signal temporal logic} (\wGSTL{}) as the weighted extension of \textit{graph-based logic} (\GL{}) which is modified from \textit{graph temporal logic} in \cite{Xu2019}. 
\vspace{-4pt}
\begin{figure}[H]
    \centering
		\includegraphics[scale=0.25]{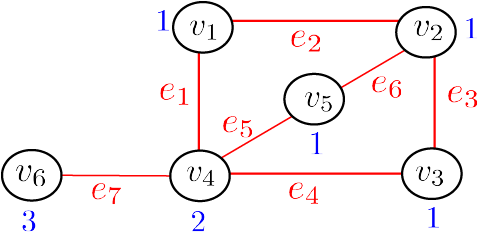}
		
		\caption{A graph-based trajectory $\graphTraj$ on an undirected graph $\graphG=(\nodeSet,E)$ with $\nodeSet=\{\node_1, \node_2, ..., \node_6\}$ and $E=\{e_1, e_2, ..., e_7\}$. Blue numbers indicate the node values at a fixed time-step $\timeIndex$. }
		\label{graphEx.}
\end{figure}
 \vspace{-10pt}
 \subsection{Graph-Based Logic}\label{SubSec:GL}

In this subsection, we define the syntax and Boolean semantics of \textit{graph-based logic} (\GL{}) formulas. {In \GL{}, we encode different locations in a graph-structured dataset as nodes in a graph $G(V,E)$, and we use the edges to encode the neighbor connections of a location demonstrated as a node.} For \GL{} formulas, we define $\neighbor\node$ to denote the set of neighbors of a node $\node$, where the subscript N stands for ``neighbor". The number of the neighbors of the node $\node$ is denoted by $|\neighbor\node|$.
 We define the syntax of \GL{} formulas as follows. 
\begin{align}
    \formulaG:=\top|\pi|\lnot\formulaG|\formulaG^1\land\formulaG^2|\Gforall{\neighbor\formulaG}|\Gexists{\neighbor\formulaG},
\end{align}

\noindent{where} $\top$ stands for the Boolean constant $\ltrue$,
$\pi$ is an atomic predicate in the form of an inequality  $\STLf(\graphTraj(\node, \timeIndex))>0$ in the form $\STLf(\graphTraj(\node,\timeIndex))=\STLCo^T\graphTraj(\node,\timeIndex)-\STLbias$,  $\STLCo\in\real^{\dimcounter}$, and  $\STLbias\in \real$; $\lnot$ (negation) and $\land$ (conjunction)
are standard Boolean connectives; $\Gforall$ is a \GL{} operator called \textit{graph-based universal quantifier} and $\Gforall{\neighbor\formulaG}$ reads as ``all the neighbors of the current node satisfy $\formulaG$"; $\Gexists$ is a \GL{} operator called \textit{graph-based existential quantifier} and  $\Gexists{\neighbor\formulaG}$ reads as ``there exists at least one neighbor of the current node that satisfies $\formulaG$". We define the Boolean semantics of \GL{} formulas as follows.\par
\begin{align*}
	(\graphTraj,\node,\timeIndex)\models\pi~\mbox{iff}&~~
	\STLf(\graphTraj(\node, \timeIndex))>0\\
	(\graphTraj,\node,\timeIndex)\models\lnot\formulaG~\mbox{iff}&~~(\graphTraj,\node,\timeIndex)\not\models\formulaG,\\
	(\graphTraj,\node,\timeIndex)\models\formulaGSt\wedge\formulaGNd~\mbox{iff}&~~(\graphTraj,\node,\timeIndex)\models\formulaGSt~\mbox{and}\\
	&~~(\graphTraj,\node,\timeIndex)\models\formulaGNd,\\
	(\graphTraj,\node,\timeIndex)\models\Gexists\neighbor\formulaG~\mbox{iff}
	 &~~\exists\hatnode\in\neighbor\node~\mbox{s.t.}~(\graphTraj,\hatnode,\timeIndex)\models\formulaG,\\
	 	(\graphTraj,\node,\timeIndex)\models\Gforall\neighbor\formulaG~\mbox{iff}
	 &~~\forall\hatnode\in\neighbor\node~\mbox{s.t.}~(\graphTraj,\hatnode,\timeIndex)\models\formulaG.
\end{align*}


 The quantitative satisfaction of graph-based logic formulas at node $\node$ and at time-step $\timeIndex$  is defined as follows. 
\begin{align*}
    \begin{split}
    \robustnessG(\graphTraj,\node,\pi,\timeIndex)  &= \STLf(\graphTraj(\node, \timeIndex)),\\
    \robustnessG(\graphTraj, \node,\lnot\formulaG,\timeIndex)  &= -\robustnessG(\graphTraj, \node,\formulaG,\timeIndex),\\
    \robustnessG(\graphTraj, \node,\formulaGSt\wedge\formulaGNd,\timeIndex)  &= \min(\robustnessG(\graphTraj, \node,\formulaGSt,\timeIndex),\robustnessG(\graphTraj, \node,\formulaGNd,\timeIndex)),\\
    \robustnessG(\graphTraj, \node,\Gforall\neighbor\formulaG,\timeIndex)  &= \min\limits_{\hatnode\in\neighbor\node}{\robustnessG(\graphTraj,\hatnode,\formulaG,\timeIndex)},\\
    \robustnessG(\graphTraj, \node,\Gexists\neighbor\formulaG,\timeIndex)  &= \max\limits_{\hatnode\in\neighbor\node}\robustnessG(\graphTraj,\hatnode,{\formulaG},\timeIndex).\\
    \end{split}
\end{align*}


\subsection{Graph-Based Signal Temporal Logic}\label{SubSec:GSTL}
 The syntax of \GSTL{} formula $\formula$ is defined recursively as follows.
\begin{align*}
    \formula &:=
    \formulaG
    \mid \lnot\formula
    \mid \formula^{1}\land\formula^{2}
    \mid \lglobally_{I}\formula
    \mid \leventually_{I}\formula,
\end{align*}

\noindent{where} $\formulaG$ is a \GL{} formula, $\lnot$ (negation) and $\land$ (conjunction)
are standard Boolean connectives, $\lglobally_{I}$ is the temporal operator ``always'', and $\leventually_{I}$ is the temporal operator ``eventually''. The Boolean semantics of \GSTL{} is based on the Boolean semantics of STL \cite{Donze} and is evaluated using graph-based trajectories. The Boolean semantics of $\formulaG$ is as described in Subsection \ref{SubSec:GL}. 

\begin{example}
	In Figure \ref{graphEx.}, $\neighbor\node_4=\{\node_1, \node_3, \node_5, \node_6\}$, and graph-based trajectory $\graphTraj$ satisfies the \GL{} formula $\Gexists\neighbor(\graphTraj>2)$ only at node $\node_4$. For the time interval $I=[1,3]$, if the node value of node $\node_6$ stays greater than 2 in the time interval $\timeIndex+I$, then the \GSTL{} formula $\lglobally_{[1,3]}(\Gexists\neighbor(\graphTraj>2))$ is satisfied by graph-based trajectory $\graphTraj$ at node $\node_4$.
\end{example}
	
\subsection{Weighted Graph-Based Signal Temporal Logic}
An extension of STL is \textit{weighted STL} (wSTL), where we assign a weight to each subformula of an wSTL formula based on its importance \cite{Mehdipour2021}\cite{Yan2021}. We refer to these weights as \textit{importance weights}. In this paper, we extend wSTL to \textit{weighted} \GSTL{} (\wGSTL{}). In \wGSTL{}, in addition to defining importance weights for the subformulas, we define the importance weights for both the temporal operators and the \GL{} operators. In other words, we assign an importance weight to each time-step $\timeIndex\in{I}$, and we assign an importance weight to each neighbor node $\hatnode\in\neighbor\node$ of a node $\node$.  We define the syntax of \wGSTL{} formulas as follows.

\begin{align*}
    \wSTL := \GwSTL[\weight]
    \mid \lnot\wSTL
    \mid \wSTLstyle[\weightSt][1]\land\wSTLstyle[\weightNd][2]
    \mid 
    \walways[\Capweight][\I]\wSTL
    \mid \wfinally[\Capweight][\I]\wSTL,
\end{align*}
\noindent{where} 
$\weightSt$ and $\weightNd$ are positive importance weights on $\wSTLstyle[\weightSt][1]$ and $\wSTLstyle[\weightSt][2]$, respectively; $\Capweight=[\weight[{k_1+1}],\weight[{k_1+2}],..,\weight[{k_2}]]^{T}\in{\real^{\timeIndexNd-\timeIndexSt+1}}$ assigns a positive weight $\weight[{k'}]$ to $k'\in[\timeIndexSt,\timeIndexNd]$ in the temporal operators; $\GCapweight=[\weight[1],\weight[2],..,\weight[{\neighborI}]]^{T}\in{\real^{\neighborI}}$ assigns a positive weight to each $\hatnode\in\neighbor\node$. In the syntax of \wGSTL{} formulas, $\GwSTL[\weight]$ is defined as follows. 
\begin{align*}
	\GwSTL[\weight] :=
	\top
	\mid \pi
	\mid \lnot\GwSTL[\weight]
	\mid \wSTLstyle[\weightSt][1][\mathcal{G}]\land\wSTLstyle[\weightNd][2][\mathcal{G}]
	\mid  \Gforall{\neighbor^{\GCapweight}\GwSTL}
	\\\mid \Gexists{\neighbor^{\GCapweight}\GwSTL}
\end{align*}
\section{Weighted Graph-Based Signal Temporal Logic and Neural Networks}
In this section, we formalize the problem of classifying graph-based trajectories by inferring \wGSTL{} formulas using neural networks. We denote a set of labeled graph-based trajectories by $\Sample=\{(\graphTraj[{\counteri}], \class[{\counteri}])\}^{N_{\Sample}}_{\counteri=1}$, and a time horizon by $H=[0, h]$ (where $h\in\mathbb{T}$). We assume that the set $\Sample$ is composed of two subsets: positive subset $\PosSet=\{\graphTraj[\counteri]|\class[\counteri]=\posclass\}$ which contains the graph-based trajectories representing the desired behavior, and negative subset $\NegSet=\{\graphTraj[\counteri]|\class[\counteri]=\negclass\}$ which contains the graph-based trajectories representing the undesired behavior. For the cardinality of the defined sets, we have $\abs{\PosSet}=\PosSetSize$, $\abs{\NegSet}=\NegSetSize$, and $\TotSetSize=\PosSetSize+\NegSetSize$. \\
Inspired by \cite{Kong2017}, we define the following.\par
\begin{definition}\label{Template}
    We define a \wGSTL{} formula structure,  denoted by $\FormulaStruct$, as a \wGSTL{} formula in which the importance weights of the subformulas and the variables of the atomic predicates, and the importance weights of the \GL{} operators and the temporal operators are replaced by free parameters. In this structure, we assume that we always have at least one temporal operator and one \GL{} operator.
    
\end{definition}



\begin{definition}\label{FormulaStruct}
	We define a \StructName{} $\FormulaStruct[f]$ as a flexible extension of \wGSTL{} formula structure such that the types of the temporal operators and the types of the \GL{} operators are to be inferred from data; but the types of the Boolean connectives in the structure are fixed. In this structure, we represent the set of \GL{} operators as a set $\pGL=\{\Gforall, \Gexists\}$ from which the proper operator is to be inferred from data. Similarly, we represent the set of temporal operators as a set $\pTemp=\{\leventually^{\Capweight}_{I}, \lglobally^{\Capweight}_{I}\}$. 
\end{definition}	
\begin{example}
	In the \StructName{} $\FormulaStruct[f]:=\textsuperscript{\weight[1]}(\pTemp^{\Capweight_1}(\pGL\neighbor^{{\GCapweight}}\pi_{1}))\lor$ $\textsuperscript{\weight[2]}(\pTemp^{\Capweight_2}(\pGL{\neighbor^{\GCapweight}\pi_{2}}))$, the types of the temporal operators and the \GL{} operators, the importance weights of the subformulas and the variables of the atomic predicates $\pi_1$ and $\pi_2$, and the importance weights of the \GL{} operators and the temporal operators are to be inferred from data, but the Boolean connective $\lor$ is fixed.
\end{example}\par

After determining the proper \wGSTL{} operators in a given $\FormulaStruct[f]$, we obtain a \wGSTL{} formula $\wSTL$ that is consistent with $\FormulaStruct[f]$.\par
{In order to define the problem statement, we define the classification accuracy, denoted by $\accu$, as $\accu={\frac{\Predicted}{N_{\mathcal{D}}}}\times{100}$, where $\Predicted\in\mathbb{N}$ is the number of correctly classified graph-based trajectories $\graphTraj[i]$}.
\vspace{-5pt}

\begin{problem}\label{Prob:InferWGSTL}
	Given a set of labeled graph-based trajectories $\Sample=\{(\graphTraj[{\counteri}], \class[{\counteri}])\}^{N_{\Sample}}_{\counteri=1}$ and a \StructName{}  $\FormulaStruct[f]$, infer a \wGSTL{} formula  $\wSTL$ (i.e., select the \wGSTL{} operators and compute the parameters of that structure) to classify $\Sample$ such that the classification accuracy {$\accu$} is maximized.  
\end{problem}\par
\vspace{-5pt}
In order to solve Problem \ref{Prob:InferWGSTL}, we introduce \wGSTL{} neural networks (\wtsGLNN{}) which combines the characteristics of \wGSTL{} and neural networks. In the first step, we construct and train a \wtsGLNN{} to learn the proper \wGSTL{} operators in the \StructName{} $\FormulaStruct[f]$. In the second step, we construct and train another \wtsGLNN{} to learn the parameters in $\FormulaStruct[f]$. In \wtsGLNN{}, we combine the activation functions in a neural network with the quantitative satisfaction of \wGSTL{}. We define the quantitative satisfaction of a \wGSTL{} formula $\wSTL$ as follows.
\begin{align*}
\begin{split}
	\wSat{\graphTraj,\node}{\pi}{\timeIndex}  &= \STLf(\graphTraj(\node,\counterk)),\\
		\wSat{\graphTraj,\node}{\lnot\wSTL}{\timeIndex}  &= -\wSat{\graphTraj,\node}{\wSTL}{\timeIndex},\\
		\wSat{\graphTraj,\node}{{\wSTLstyle[\weightSt][1]}\land{\wSTLstyle[\weightNd][2]}}{\timeIndex}  &= \actFunAnd({\intervalStyle{\weight[{\counteri}]}{\wSat{\graphTraj,\node}{\wSTL^{\counteri}}{\timeIndex}}}_{\counteriequal}),\\
		\wSat{\graphTraj,\node}{{\wSTLstyle[\weightSt][1]}\lor{\wSTLstyle[\weightNd][2]}}{\timeIndex}  &= \actFunOr({\intervalStyle{\weight[{\counteri}]}{\wSat{\graphTraj,\node}{\wSTL^{\counteri}}{\timeIndex}}}_{\counteriequal}),\\
		\wSat{\graphTraj,\node}{{\walways[\Capweight][I]}\wSTL}{\timeIndex}  &= \actFunGlo(\Capweight,\{\wSat{\graphTraj,\node}{\wSTL}{\counteri}\}_{\counteriin}),\\
		\wSat{\graphTraj,\node}{{\wfinally[\Capweight][I]}\wSTL}{\timeIndex}  &= \actFunEve(\Capweight,\{\wSat{\graphTraj,\node}{\wSTL}{\counteri}\}_{\counteriin}),\\
		\wSat{\graphTraj, \node}{\Gforall\neighbor^{\GCapweight}\GwSTL}{\timeIndex}  &= \actFunGforall(\GCapweight,\{\wSat{\graphTraj,\hatnode}{\GwSTL}{\timeIndex}\\
	&~~~~~~~~~~~~~~~~~~~~~~\}_{\hatnode\in\neighbor\node}),\\
		\wSat{\graphTraj, \node}{\Gexists\neighbor^{\GCapweight}\GwSTL} {\timeIndex} &= \actFunGexists(\GCapweight,\{\wSat{\graphTraj,\hatnode}{\GwSTL}{\timeIndex}\\
	&~~~~~~~~~~~~~~~~~~~~~~\}_{{\hatnode\in\neighbor\node}}),
		\end{split}
\end{align*}

\noindent{where} $\actFunAnd,\actFunOr: \real^{2}_{>0}\times\real^{2}\limplies\real$, $\actFunGlo,\actFunEve: \real^{\timeIndexNd-\timeIndexSt+1}_{>0}\times\real^{\timeIndexNd-\timeIndexSt+1}\limplies\real$, and $\actFunGforall,\actFunGexists: \real^{\abs{\neighbor\node}}_{>0}\times\real^{\abs{\neighbor\node}}\limplies\real$ are activation functions corresponding to $\land$, $\lor$, $\lglobally^{\Capweight}_{I}$, $\leventually^{\Capweight}_{I}$, $\Gforall$, and $\Gexists$ operators, respectively. We denote an activation function with $\actFun[\counteri]$, where $\counteri\in{\mathcal{A}}=\{\land,
\lor,\lglobally,\leventually,\Gforall,\Gexists\}$. 

For defining the activation functions, we use the variable $\promVar$, where $\promVar$ is a positive real number. We define the activation function corresponding to each operator in the set $\mathcal{A}$ as follows. 
\vspace{-10pt}
\begin{align}\label{activeFuncs}
\begin{split}
    \actFunAnd({\intervalStyle{\weight[{\counteri}]}{\wSat{\graphTraj,\node}{\wSTL^{\counteri}}{\timeIndex}}}_{\counteriequal},\promVar)  &= \frac{\lSigma{2}{\iprim=1}{\normalweight[\iprim]\ActiveS_{\iprim}\robustness[\iprim]}}{\lSigma{2}{\iprim=1}{\normalweight[\iprim]\ActiveS_{\iprim}}},\\
    \actFunOr({\intervalStyle{\weight[{\counteri}]}{\wSat{\graphTraj,\node}{\wSTL^{\counteri}}{\timeIndex}}}_{\counteriequal},\promVar)  &= -\frac{\lSigma{2}{\iprim=1}{\normalweight[\iprim]\ActiveS_{\iprim}\robustness[\iprim]}}{\lSigma{2}{\iprim=1}{\normalweight[\iprim]\ActiveS_{\iprim}}},\\
 \actFunGlo(\Capweight,\{\wSat{\graphTraj,\node}{\wSTL}{\counteri}\}_{\counteriin},\promVar)  &= 
     \frac{\lSigma{\timeIndexNd}{\iprim=\timeIndexSt}{\normalweight[\iprim]\ActiveS_{\iprim}\robustness[\iprim]}}{\lSigma{\timeIndexNd}{\iprim
     		=\timeIndexSt}{\normalweight[\iprim]\ActiveS_{\iprim}}},\\
      \actFunEve(\Capweight,\{\wSat{\graphTraj,\node}{\wSTL}{\counteri}\}_{\counteriin},\promVar)  &= 
     \frac{\lSigma{\timeIndexNd}{\iprim=\timeIndexSt}{\normalweight[\iprim]\ActiveS_{\iprim}\robustness[\iprim]}}{-\lSigma{\timeIndexNd}{\iprim=\timeIndexSt}{\normalweight[\iprim]\ActiveS_{\iprim}}},\\
     \actFunGforall(\GCapweight,\{\wSat{\graphTraj,\hatnode}{\GwSTL}{\timeIndex}\}_{\hatnode\in\neighbor\node},\promVar)  &= 
     \frac{\lSigma{\neighborI}{\iprim=1}{\normalweight[\iprim]\ActiveS_{\iprim}\robustness[\iprim]}}{\lSigma{\neighborI}{\iprim=1}{\normalweight[\iprim]\ActiveS_{\iprim}}}\\
     \actFunGexists(\GCapweight,\{\wSat{\graphTraj, \hatnode}{\GwSTL}{\timeIndex}\}_{\hatnode\in\neighbor\node},\promVar)  &= 
     \frac{\lSigma{\neighborI}{\iprim=1}{\normalweight[\iprim]\ActiveS_{\iprim}\robustness[\iprim]}}{-\lSigma{\neighborI}{\iprim=1}{\normalweight[\iprim]\ActiveS_{\iprim}}},\\
\end{split}
\end{align}
\noindent{where} $\normalweight[\iprim]$ is the normalized weight and $\timeInterval$; in the activation functions of $\land$ and $\lor$ Boolean connectives, $\normalweight[\iprim]=\weight[{\iprim}]/\lSigma{2}{\counterj=1}{\weight[{\counterj}]}$; In the activation functions of $\lglobally^{\Capweight}_{I}$ and $\leventually^{\Capweight}_{I}$ operators, $\normalweight[\iprim]=\weight[{\iprim}]/\lSigma{\timeIndexNd}{\counterj=\timeIndexSt}{\weight[{\counterj}]}$; in $\Gforall$ and $\Gexists$ graph operators, $\normalweight[\iprim]=\weight[{\iprim}]/\lSigma{\neighborI}{\counterj=1}{\weight[{\counterj}]}$. {$\normalweight[m]$ aims to normalize the importance weight of the $m$-th subformula in $\wSTL$ to the range of $[0,1]$ such that $\overline{w}_m$ can reflect the
importance of the $m$-th subformula in determining the quantitative satisfaction of the
whole \wGSTL{} formula $\wSTL$.} 
In the activation function of $\land$  operator, $\robustness[\iprim]=\wSat{\graphTraj, \node}{\wSTL^{\iprim}}{\timeIndex}$; in the activation function of $\lor$ operator,  $\robustness[\iprim]=-\wSat{\graphTraj, \node}{\wSTL^{\iprim}}{\timeIndex}$; in the activation function of $\Gforall$ operator, $\robustness[\iprim]=\wSat{\graphTraj,\hatnode}{\GwSTL}{\timeIndex}$; in the activation function of $\Gexists$ operator, $\robustness[\iprim]=-\wSat{\graphTraj,\hatnode}{\GwSTL}{\timeIndex}$; in the activation function of $\lglobally^{\Capweight}_{I}$ operator, $\robustness[\iprim]=\wSat{\graphTraj,\node}{\wSTL}{\iprim}$;
 and in the activation function of $\leventually^{\Capweight}_{I}$ operator, $\robustness[\iprim]=-\wSat{\graphTraj, \node}{\wSTL}{\iprim}$;
  in the activation functions of $\land$ and $\lor$ operators, $\siBool$; in the activation functions of $\lglobally^{\Capweight}_{I}$ and $\leventually^{\Capweight}_{I}$ operators, $\siTemp$; in the activation functions of $\Gforall$ and $\Gexists$ operators, $\siGraph$.\par

\section{Methodology}\label{Methodology}

In this section, we introduce a framework and algorithms to solve Problem \ref{Prob:InferWGSTL} for a given \StructName{} $\FormulaStruct[f]$ with $M$ undetermined temporal operators and $L$ undetermined \GL{} operators using \wtsGLNN{}. For the training of \wtsGLNN{}, we design a loss function to meet the following two requirements: 1) the loss should be small when the inferred formula is satisfied by graph-based trajectories in $\Sample[P]$ and violated by the graph-based trajectories in $\Sample[N]$; 2) the loss should be large when the inferred formula is not satisfied by the graph-based trajectories in $\Sample[P]$ and not violated by the graph-based trajectories in $\Sample[N]$. We define the loss function as follows.
\begin{equation}\label{LossFun}
	\LossFun(\wSTL)=\lSigmaTop{\TotSetSize}{\counteri=1}{e^{-\tuningParam\class[\counteri]\wSatTwo{\graphTraj[\counteri]}{\wSTL}}},
\end{equation}
\noindent{where} $\graphTraj[\counteri]$ denotes the $\counteri$-th graph-based trajectory in $\Sample$, and $\tuningParam>0$ is a tuning parameter. $\LossFun(\wSTL)$ is small for the cases where $\Setlabel\wSatTwo{\graphTraj[\counteri]}{\wSTL}>0$ and increases exponentially when $\Setlabel\wSatTwo{\graphTraj[\counteri]}{\wSTL}<0$.\par
\begin{algorithm}
	\small
	\DontPrintSemicolon
	
	\Input{a set of labeled graph-based trajectories $\Sample$\newline
    \wGSTL~ structure $\FormulaStruct[f]$\newline
    number of iterations $\batchSize$\newline
    number of subformulas $\SubFormNo$
    } 
  \Output{Learned \wGSTL{} formula $\wSTL$}

  Construct a \wtsGLNN{} based on the given structure $\FormulaStruct[f]$\;
  
  Initialize $\boldsymbol{w}^{\FormulaStruct[f]}$, $\STLCo$ and $\STLbias$\;
  
  Initialize $\{b_{i}\}^{M}_{i=1}$, $\{b'_{i'}\}^{L}_{i=1}$, respectively corresponding to $\pTemp$ and $\pGL$\;
 
  $\{b_{i}\}^{M}_{i=1},\{b'_{i'}\}^{L}_{i'=1}\gets$\wGSTL-NN($\Sample, \FormulaStruct[f], \batchSize, \SubFormNo,\{\{b_{i}\}^{M}_{i=1},\{b'_{i'}\}^{L}_{i'=1}\})$\;
  {\For{$i=1,...,M$}
    	{
  	
  	\lIf{$b_i\geq{0}$}{$b_i\gets{\globallyCo}$}
  	\lElse{$b_i\gets{\eventuallyCo}$}
  }
  }
  {\For{$i'=1,...,L$}
  {
  	
  	\lIf{$b'_{i'}\geq{0}$}{$b'_{i'}\gets{\GforallCo}$}
  	\lElse{$b'_{i'}\gets{\GexistsCo}$}
  }
  }

$\FormulaStruct\gets$ Selected operators from the $\pGL$ and the $\pTemp$\

 $\boldsymbol{w}^{\FormulaStruct[f]},\STLCo,\STLbias\gets$\wGSTL-NN($\Sample, \FormulaStruct, \batchSize, \SubFormNo,\{\boldsymbol{w}^{\FormulaStruct[f]},\STLCo,\STLbias\})$\;
 
 $\wSTL\gets{\boldsymbol{w}}^{\FormulaStruct[f]},\STLCo,\STLbias$
  
\Return{\wSTL}
	\caption{Two-step procedure of inferring a \wGSTL{} formula}
    \label{alg:w-GSTL}

\end{algorithm}

We compute a \wGSTL{} formula in two steps: 1) determining the proper temporal and \GL{} operators in the given \StructName{} $\FormulaStruct[f]$; 2) learning the parameters of the \StructName{} $\FormulaStruct[f]$. For each step, we construct and train a separate \wtsGLNN{}.\par

Algorithm \ref{alg:w-GSTL} illustrates the two-step procedure of learning a \wGSTL{} formula from a given set of labeled graph-based trajectories $\Sample$ and a given \StructName{} $\FormulaStruct[f]$. 
\paragraph{Step 1} In step 1, we initialize two sets of coefficients:  (a) $\{b_{i}\}^{M}_{i=1}$ corresponding to $M$ undetermined temporal operators; (b) $\{b'_{i'}\}^{L}_{i'=1}$ corresponding to $L$ undetermined \GL{}  operators, where $b_i,b'_{i'}\in\real$ (Line 3 in Alg. \ref{alg:w-GSTL}). In this step, for the activation functions of $\lglobally^{\Capweight}_{I}$ and $\leventually^{\Capweight}_{I}$, we define the coefficients $\globallyCo=+1$ and $\eventuallyCo=-1$, respectively. Similarly, for the activation functions of $\Gforall$ and $\Gexists$, we define the coefficients $\GforallCo=+1$ $\GexistsCo=-1$, respectively. We construct and train a \wtsGLNN{} (demonstrated in Alg. \ref{alg:general-GSTL}) to learn $\{b_{i}\}^{M}_{i=1}$ and $\{b'_{i'}\}^{L}_{i'=1}$ (Line 4 in Alg. \ref{alg:w-GSTL}). For each undetermined temporal operator, the sign of the returned $b_i$ determines the proper selection from the set $\pTemp$. In other words, if $b_i\geq{0}$, then we have $b_i=\globallyCo$. This means the proper selection for the temporal operator corresponding to $b_i$ is $\lglobally^{\Capweight}_{I}$. If $b_i<{0}$, then $\leventually^{\Capweight}_{I}$ is the proper selection from the set $\pTemp$ (Line 5 to 8 in Alg. \ref{alg:w-GSTL}). Similarly, for each undetermined \GL{} operator, the sign of the returned $b'_{i'}$ determines the proper selection of the undetermined \GL{} operator from the set $\pGL$ (Lines 9 to 12 in Alg. \ref{alg:w-GSTL}).
\begin{algorithm}
 \scriptsize
	\small
	\DontPrintSemicolon
	
	\Input{a set of labeled graph-based trajectories $\Sample$\newline
		\StructName{} structure $\FormulaStruct[f]$ (or a formula structure $\FormulaStruct$)\newline
		number of iterations $\batchSize$\newline
		number of subformulas $\SubFormNo$\
		
	}
	\Output{\param{: $\{b_{i}\}^{M}_{i=1}$ and $\{b_{i'}\}^{L}_{i'=1}$, and the parameters of $\FormulaStruct[f]$}}
	
	Construct a \wtsGLNN{} based on the given structure $\FormulaStruct[f]$\;
	
	
	{
	
	\For{$\counterk=$1, 2, ..., $\batchSize$}{Select a mini-batch data $\Sample[\counterk]$ from $\Sample$
		
		\For{$\counterj=$ 1, 2, ..., $\SubFormNo$}{

			
		$\param\gets{\forward(\Sample[\counterk],\subformulas{j},\actFun[j],\param)}$\;} 
		
		
		Compute $\LossFun(\wSTL)$ using \eqref{LossFun}\;
		
		
		$\param\mbox{~and~}\wSatTwo{\graphTraj[\counteri]}{\wSTL}\gets{\back(\Sample[\counterk],\param, {\LossFun(\wSTL)})}$\

	}
	
    }
	\caption{\wGSTL-NN: neural network for learning \wGSTL{} formulas}
	\label{alg:general-GSTL}
\Return{\param}	
\end{algorithm}

\paragraph{Step 2}After determining the proper operators, we construct and train another \wtsGLNN{} (demonstrated in Alg. \ref{alg:general-GSTL}) to learn parameters of the \StructName{} $\FormulaStruct[f]$ including $\STLCo$, $\STLbias$, and all the importance weights in the \StructName{} $\FormulaStruct[f]$ that we denote them by $\boldsymbol{w}^{\FormulaStruct[f]}$ (Lines 13 to 16 in Alg. \ref{alg:general-GSTL}). \par

Algorithm \ref{alg:general-GSTL} illustrates \wtsGLNN{} that we use to learn \wGSTL{} formulas. In Algorithm \ref{alg:general-GSTL}, $\param$  denotes a set of parameters that we calculate at each step of the two-step process of learning a \wGSTL{} formula. In step 1, $\param$ includes $\{b_{i}\}^{M}_{i=1}$ and $\{b_{i'}\}^{L}_{i'=1}$. In step 2, $\param$ includes the parameters of the \StructName{} $\FormulaStruct[f]$ (including $\STLCo$, $\STLbias$, and $\boldsymbol{w}^{\FormulaStruct[f]}$). In Algorithm \ref{alg:general-GSTL}, we denote the forward-propagation operation by $\forward(\Sample[\counterk],\subformulas{j},\actFun[\counteri],\param)$, where $\Sample[k]$ denotes the selected mini-batch  from the set $\Sample$, $\subformulas{\counterj}$ is the $j$th subformula in $\FormulaStruct[f]$, $\actFun[j]$ is activation function  corresponding to the \wGSTL{} operator or Boolean connective in $\subformulas{\counterj}$, $\param$ is to be calculated, and $\wSatTwo{\graphTraj[\counteri]}{\wSTL}$ is the output of forward-propagation (Line 5 in Alg. \ref{alg:general-GSTL}). More clearly, after determining the \wGSTL{} operators, \wtsGLNN{} calculates the quantitative satisfaction of the inferred $\wSTL$ with respect to $\graphTraj[i]$ through forward-propagation.  Also, We denote the back-propagation operation by $\back(\Sample[\counterk],\param,\LossFun(\wSTL))$, where $\param$ is to be updated (Line 8 in Alg. \ref{alg:general-GSTL}). The proposed algorithms are implemented in a Python toolbox\footnote{https://github.com/kazhirota7/Graph-based-wSTLNN}. \par

{The time complexity for learning a \wtsGLNN{} for a given graph $G(V,E)$ is $\mathcal{O}({u_1}{u_2}{u_3}{n_E}{u_4})$, where $u_1$ is the number of training samples, $u_2$ is the number of graph-based trajectories in each sample, $u_3$ is the number of epochs, $n_E$ is the number of nodes in the graph $G(V,E)$, and $u_4$ is the total number of temporal and graph-based operators in the \StructName{} $\FormulaStruct[f]$. }



\section{Case Studies}
In this section, we assess the performance of \wtsGLNN{}. We first use a meteorological dataset in Australia to predict rainfall. Then, we predict the severity of lockdown measures using COVID-19 data in Italy. The performance of \wtsGLNN{} is compared with some other standard classification algorithms. The \StructName{} that we use for these two case studies is  $\FormulaStruct[f]:=\textsuperscript{\weight[1]}(\pTemp^{\Capweight_1}(\pGL\neighbor^{{\GCapweight}}\pi_{1}))\lor(\lnot\textsuperscript{\weight[2]}($ $(\pTemp^{\Capweight_2}\pGL{\neighbor^{\GCapweight}\pi_{2}})))$.  {For both case studies, we use  the Adam optimization algorithm \cite{KingmaB14} to learn the neural network.} {The $\STLbias$\ parameter is initialized as 0 to prevent initial bias. Furthermore, all of the importance weights are initialized at the same value of 0.5, also to prevent initial bias.}

\subsection{Case Study I: Rain Prediction}

In this subsection, we use \wtsGLNN{} to predict rainfall in regions of Australia. The dataset is acquired from the Australian Government Bureau of Meteorology\footnote{http://www.bom.gov.au/climate/data/}\cite{Young2017}. The dataset that we use is composed of weather-related data in 49 regions of Australia measured daily from March 1st, 2013 to June 25th, 2017, including minimum 
sunshine ($\graphTraj^{1}$),
cloud at 9am ($\graphTraj^{2}$), cloud at 3pm ($\graphTraj^{3}$), 
and whether there was any rain on the given day (-1 or 1) ($\graphTraj^{4}$). {Only the most important attributes are mentioned due to space limitations}. The criterion for the binary classification is the rain prediction for the following day (-1 for no rain, 1 for rain). We construct a graph structure of the Australian regions, considering any region within a 300 km radius of another region to be neighbors. {A separate neural network is learned for each region and its neighbors to prevent scalability issues when broader regions are considered.} We utilize the zero imputation method for missing values in the input data. The dataset is divided into a proportion of 0.8:0.2 for the training dataset and testing dataset, respectively, resulting in 1262 total data points per region for the training dataset and 316 total data points per region for the testing dataset.\par

{For a demonstration of how the importance weights and predicates can be interpreted, we consider the learned neural network for Albury and its neighbors: Wagga Wagga, Canberra, Tuggeranong, Mount Ginini, Bendigo, Sale, Melbourne Airport, Melbourne, and Watsonia.} We set $K_I$ consecutive days of data as one instance of the dataset in our experiment. The dataset is passed through two separate neural networks. In the first step, we determine the temporal operators and the \GL{} operators to be applied on $I_i$ and $I_{ii}$ intervals, respectively. In the second step, we learn the parameters of the \StructName{} $\FormulaStruct[f]$. \par
In the experiment, we set $K_I$ = 15, $I_i$ = [0, 6], $I_{ii}$ = [7, 14], \tuningParam{} = 1, and \promVar{} = 1. The learned \wGSTL{} formula is 
	$\wSTL=\textsuperscript{\weight[1]}(\walways[\Capweight_1][[0,6]](\Gexists{\neighbor}^{\GCapweight}\pi))\lor 
	(\lnot\textsuperscript{\weight[2]}(\wfinally[\Capweight_2][[7,14]](\Gexists{\neighbor^{\GCapweight}}\pi)))$, and the inferred predicate $\pi$ is $\pi :=  - 0.0298\graphTraj^{1} + 0.0226\graphTraj^{2}
   + 0.0222\graphTraj^{3}-0.0309\graphTraj^{4} \leq 0.6593)$, the importance weights for $I_{i}$ are\footnote{The importance weights for $I_{ii}$ were omitted due to space limitation.} $\Capweight_1 = [0.1087, 0.2210, 0.0655, 0.1927,
   0.0163, 0.1349, 0.2609]$, the normalized importance weights for Wagga Wagga, Canberra, Tuggeranong, Mount Ginini, Bendigo, Sale, Melbourne Airport, Melbourne, and Watsonia, respectively, are $\GCapweight = [0.0443, 0.1439, 0.0319, 0.1930, 0.1299,
   0.0000, 0.1984,\\ 0.1719, 0.0867]$, and the normalized importance weights for the $\lor{}$ operator are: $\weight[1]$ = 0.6891, $\weight[2]$ = 0.3109.\par

We evaluate the performance of the proposed algorithm by applying some standard classification methods such as K-nearest neighbors (KNN) and decision tree (DT), {kernel method such as support vector machine (SVM)}, and an artificial neural network (ANN) algorithm on the same dataset. 

\begin{table}[hbt!]
\addtolength{\tabcolsep}{-5pt}
\begin{center}
\begin{tabular}{| c c |}
\hline
 \textbf{Method} & \textbf{{Obtained Accuracy}} \\
 & \textbf{{for the Test Dataset (\%)}}\\
 
 \hline
 Decision Tree & 76.14 \\ 
 \hline
 K-Nearest Neighbors (KNN) & 81.04   \\
 \hline
  Support Vector Machine (SVM) & 82.61  \\
 \hline
 ANN (Sequential Model) & 84.73   \\
 \hline
 \textbf {\wtsGLNN{} (this paper)} & \textbf {81.69} \\
 \hline
\end{tabular}
\caption{{Obtained classification accuracy on the rainfall test dataset of 49 Australian regions.}}
\label{table:1}
\end{center}
\end{table}

\wtsGLNN{} produces a higher accuracy than both KNN and DT (Table \ref{table:1}). {Although SVM and ANN produce higher accuracy than \wtsGLNN{} because they are not restricted by the \wGSTL{} formula, \wtsGLNN{} can generate human-readable results, unlike the other algorithms. Using the learned \wGSTL{} formula, the learned importance weights, and the predicates, we can interpret the decision-making of the classifier, rather than merely interpreting the results generated by a black box. The signs and the magnitudes of the predicates can be used to analyze the correlation between each meteorological input and rain prediction. The greater magnitude of the predicate suggests a stronger correlation between the input and rain prediction, and the sign of the predicate indicates whether the input has a positive or a negative relationship with rain prediction.} For instance, the coefficients associated with the predicates suggest that larger amount of clouds at 9 am ($\graphTraj^{2}$), a larger amount of clouds at 3 pm($\graphTraj^{3}$), and more rainfall during the interval ($\graphTraj^{4}$) in the input data correlates to a larger probability that there will be rainfall on the target date, while the larger amount of sunshine ($\graphTraj^{1}$) correlates to a larger probability that there will not be rainfall on the target date. {Furthermore, the learned importance weights associated with \Gexists{} suggest that the meteorological conditions in Mount Ginini and Melbourne Airport are the best indicators for predicting rain in Albury. }\par
\subsection{Case Study II: Classifying COVID-19 Lockdown Measures}
In this subsection, we use simulated COVID-19 datasets of 20 Italian regions from the DiBernardo Group Repository\cite{Rossa2020}. 
The dataset is composed of a time-series dataset for each region in Italy. The inputs of each time series are  percentage of people infected ($\graphTraj^{1}$), quarantined ($\graphTraj^{2}$), deceased ($\graphTraj^{3}$), and hospitalized due to COVID-19 ($\graphTraj^{4}$). The data is simulated for the social distancing parameter of 0.3 for strict lockdown measures and 1 for no lockdown measures. Each of the inputs of the data was recorded daily for 365 days.
We turn this case study into a binary classification by labeling ``strict lockdown measures" with -1 and ``no lockdown measures" with 1.
{For the demonstration of the learned importance weights and predicates, we consider the learned spatial-temporal properties of Abruzzo and its neighboring regions: Lazio, Marche, and Molise.} \par

In this case study, we use $K_I$ = 30, $I_i$ = [0, 14], $I_{ii}$ = [15, 29], \tuningParam{} = 1, and \promVar{} = 1. We divide the dataset into 472 sets of $K_I$ time instances for training dataset and 200 sets of $K_I$ time instances for testing dataset. The \wGSTL{} formula with learned operators is $\wSTL=\textsuperscript{\weight[1]}(\walways[\Capweight_1][[0,14]](\Gexists\neighbor^{{\GCapweight}}\pi))\lor(\lnot\textsuperscript{\weight[2]}( \wfinally[\Capweight_2][[15,29]](\Gexists\neighbor^{{\GCapweight}}\pi)))$, where the inferred predicate $\pi$ is $\pi := (-4.4617\graphTraj^{1} - 4.3504\graphTraj^{2} - 3.2291\graphTraj^{3} - 4.4045\graphTraj^{4}
   \leq -0.1421)$, the normalized importance weights for $I_{ii}$ are $\Capweight_2 = [0.0006, 0.0007, 0.0008, 0.0008, 0.0007, 0.0008,
   0.0007,\\ 0.0007, 0.0007, 0.0006, 0.0007, 0.0008, 0.0007, 
   0.1663, 0.8064]$, {the} normalized importance weights for Lazio, Marche, and Molise, respectively, are $\GCapweight = [0.0002,  0.0001, 0.9997]$, and the normalized importance weights for the $\lor{}$ operator are: $\weight[1]$ = 0.9998, $\weight[2]$ = 0.0002.\par 
   {We apply the K-nearest neighbors, decision tree, and support vector machine algorithms on this dataset to conduct the binary classification.} 
The learned \wGSTL{} formula provides us with the spatial-temporal properties of the dataset and determines whether there is a strict lockdown measure in the region or not. {Furthermore, the accuracy of \wtsGLNN{} matches that of K-nearest neighbors, decision tree, and support vector machine for the COVID-19 {test} dataset, all achieving an accuracy of 100\%. 
Nevertheless, \wtsGLNN{} provides important information for analysis that would not be possible to retrieve when using other algorithms, which makes \wtsGLNN{} useful for applications when interpretation for the decision-making of computers is necessary.} \par

 \vspace{-3pt}
\section{Conclusion}
In this paper, we proposed a framework that combined neural networks and \wGSTL{} for learning spatial-temporal properties from data. The proposed approach represents the learned knowledge in a human-readable form.
As the future direction, we plan to extend this approach to scenarios where only the positive data is available. Also, we aim to apply \wtsGLNN{} in the settings of deep reinforcement learning (deep RL) to improve the interpretability of the deep RL, where we deal with graph-structured problems.

\titleformat{\section}{\centering\normalfont\scshape}{\appendixname~\thesection }{0em}{~}

\bibliographystyle{IEEEtran}
\bibliography{biblography}

\end{document}